**Title Page**

**Title: Machine learning algorithms to predict the risk of rupture of intracranial aneurysms: a systematic review.**

**Authors**: Karan Daga[1,2], Siddharth Agarwal[1], Zaeem Moti[2], Matthew BK Lee[3], Munaib Din[2], David Wood[1], Marc Modat[1], Thomas C Booth[1,4].

**Affiliations**:

[1] School of Biomedical Engineering & Imaging Sciences, King's College London, London UK.

[2] Guy's and St. Thomas' NHS Foundation Trust, London UK.

[3] University College London Hospital NHS Foundation Trust, London UK.

[4] Department of Neuroradiology, King's College Hospital, London UK.

**Corresponding Author:**

**Name**: Dr. Thomas C Booth

**Email**: thomas.booth@kcl.ac.uk

**Telephone**: +44 (0) 20 7848 9568

**Department**: School of Biomedical Engineering & Imaging Sciences

**Institute**: King's College London

**Address**: BMEIS, King's College London. 1 Lambeth Palace Road. London, UK SE1 7EU.

**Funding**: T.C.B. and D.W. are supported by the UK Medical Research Council (MR/W021684/1) and by the Wellcome EPSRC Centre for Medical Engineering at King's College London (203148/Z/16/Z)

**Competing Interests:** All authors declare no financial or non-financial competing interests.

**Data Availability:** The datasets used and/or analyzed during the current study are available from the corresponding author on reasonable request.

**Author Contributions**

1. KD: Conception and design of the review, acquisition, analysis and interpretation of the data and writing the manuscript.
2. SA: Conception and design of the review, acquisition, analysis and interpretation of the data.
3. ZM: Acquisition and analysis of the data
4. ML: Conception and design of the review, interpretation of the data.
5. MD: Conception and design of the review, acquisition of the data
6. DW: Conception and design of the review
7. MM: Conception and design of the review
8. TCB: Conception and design of the review, acquisition, analysis and interpretation of the data and writing the manuscript.

All authors have read and approved the final manuscript. TCB is the corresponding author and guarantor of the study.

**Ethics approval:**
No ethics approval was required for this systematic review of the existing literature.

**Consent:**
No consent was required for this systematic review of the existing literature.

**Abstract**

**Purpose**: Subarachnoid haemorrhage is a potentially fatal consequence of intracranial aneurysm rupture, however, it is difficult to predict if aneurysms will rupture. Prophylactic treatment of an intracranial aneurysm also involves risk, hence identifying rupture-prone aneurysms is of substantial clinical importance. This systematic review aims to evaluate the performance of machine learning algorithms for predicting intracranial aneurysm rupture risk.

**Methods**: MEDLINE, Embase, Cochrane Library and Web of Science were searched until December 2023. Studies incorporating any machine learning algorithm to predict the risk of rupture of an intracranial aneurysm were included. Risk of bias was assessed using the Prediction Model Risk of Bias Assessment Tool (PROBAST). PROSPERO registration: CRD42023452509.

**Results**: Out of 10,307 records screened, 20 studies met the eligibility criteria for this review incorporating a total of 20,286 aneurysm cases. The machine learning models gave a 0.66-0.90 range for performance accuracy. The models were compared to current clinical standards in six studies and gave mixed results. Most studies posed high or unclear risks of bias and concerns for applicability, limiting the inferences that can be drawn from them. There was insufficient homogenous data for a meta-analysis.

**Conclusions**: Machine learning can be applied to predict the risk of rupture for intracranial aneurysms. However, the evidence does not comprehensively demonstrate superiority to existing practice, limiting its role as a clinical adjunct. Further prospective multicentre studies of recent machine learning tools are needed to prove clinical validation before they are implemented in the clinic.

**Keywords**: Aneurysm; Intracranial Aneurysm; Machine Learning; Subarachnoid Haemorrhage; Cerebrovascular.

## Introduction

The prevalence of intracranial aneurysms in the general population is approximately 3.2%.[1] Rupture risk can vary greatly depending on aneurysm morphology, for example the relative risk of rupture of larger aneurysms (>15mm) is 15.4 in comparison to smaller (<5mm) aneurysms.[2] Similarly, patient demographics can play a significant role as smokers have a relative risk of rupture of 1.7 in comparison to non-smokers, and female patients have a relative risk of 2.0.[2] The management of unruptured intracranial aneurysms (UIA) is contextualised by patient age, comorbidity, clinical presentation and the estimated rupture risk. Cerebral aneurysms can be managed with surgical procedures such as microsurgical clipping, endovascular treatment (including coiling, stenting or intrasaccular device implantation), or they can be monitored with imaging without initial intervention.

It is essential to stratify rupture risk to guide appropriate management. Although no universally-agreed reference standard criteria exists for prognostication and definition, scoring systems have been devised as decision support tools. A commonly applied example is the PHASES score; 'Population, Hypertension, Age, Size of aneurysm, Earlier subarachnoid haemorrhage (SAH) from another aneurysm and Site of aneurysm'.[3] Studies assessing the PHASES score demonstrated sensitivity and specificity of 0.50 and 0.81, respectively, during a total follow-up of 3064 person-years, indicating insufficient predictive accuracy when applied independently in clinical practice.[4] Other commonly used scoring criteria are the 'Unruptured Intracranial Aneurysm Treatment Score' (UIATS), the 'International Study on Unruptured Intracranial Aneurysms' (ISUIA) score and the 'Unruptured Cerebral Aneurysm Study' (UCAS) score.[5-7] Consequently, the variety of rupture risk scoring systems has resulted in heterogeneity in clinical management planning and patient follow up. For example, UIA size is typically considered to be one of the most predictive criteria for rupture risk but is also considered by many to be inadequate in isolation given the high prevalence of rupture in aneurysms < 3 mm.[8,9] Because three-dimensional (3D) morphology is also an important risk factor, increasingly morphology has been used in decision making.[10]

Given that there remains an unmet clinical need to optimise and homogenise decision making for UIAs, and given that machine learning (ML) has shown great promise in healthcare with its capability to process large multimodal datasets (resulting in a wide range of use-cases from diagnostics to management planning), it is plausible that ML-based data-driven decision tools will be added to the armamentarium of clinicians. However, whilst several recent ML models have been developed for this purpose, there has been a lack of implementation in the clinic. This study therefore aimed to systematically review and summarize the accuracy of ML models to predict UIA rupture. The review process allowed us to explore the limitations and implications of using UIA decision support tools, and provides a baseline for future research.

## Materials and Methods

This systematic review is PROSPERO registered (International prospective register of systematic reviews) (CRD42023452509). The review followed the Preferred Reporting Items for Systematic Reviews and Meta-Analysis (PRISMA) guidelines, informed by the Checklist for Artificial Intelligence in Medical Imaging (CLAIM), and Cochrane review methodology for developing inclusion and exclusion criteria, search methodology and quality assessment.[11-14]

### Search Strategy and Selection Criteria

A ‘sensitive search’ strategy (Supplementary Material) was undertaken consisting of relevant search terms and subject headings, including exploded terms and Medical Subject Headings (MeSH) terms, without language restrictions.[13] The search was conducted in the following medical databases: EMBASE, MEDLINE, Web of Science and the Cochrane Register to include articles published until December 2023. Pre-prints, conference abstracts and non-peer reviewed articles were excluded (Fig. 1). Three reviewers (radiologists with 2, 2 and 7 years neurovascular research experience, respectively) all performed the literature search and selection; consensus was achieved with a fourth reviewer (15 years neurovascular research experience).

### Inclusion criteria

This review included primary studies incorporating any ML methodology to predict the rupture risk of UIAs. In a subset of studies that evaluated ‘stability’ of an aneurysm (as opposed to rupture), the authors defined stability as a composite outcome which included rupture as well as aneurysm growth and/or presence of symptoms.

### Exclusion criteria

Studies were excluded if epidemiological and assessing individual risk factors without developing a predictive model, or developing models that only predicted current rupture status (i.e., models performing binary classification of aneurysms as 'ruptured' or 'unruptured'). Animal studies and studies in paediatric populations (age < 18 years) were also excluded.

Index test and reference standard

The index test was the ML model predicting rupture risk of an aneurysm. In most studies, the reference standard was either the event of rupture or 'stability' during follow up. A few studies used an established prediction model (e.g., PHASES) as the reference standard, specified in Tables 1 and 2.

Data extraction and analysis

We extracted data on imaging modality used; reference standard employed; period of risk assessment/follow-up; index test ML model; model features (grouped as clinical, morphological or hemodynamic features); study population; datasets (training/validation/testing); and inclusion/exclusion criteria. We recorded information on whether testing data sets were used for either internal or external validation. Internal validation consists of testing a model on any form of internal 'hold-out' data including temporally-spaced datasets from the same institution. External validation consists of testing a model using data that was separated geographically (different institutions) from the training datasets. The performance accuracy of the best performing ML model or composite ML model were obtained, with the majority presented as receiver operating characteristic area under the curve (ROC-AUC); where possible we also calculated other measures of accuracy including sensitivity, specificity, balanced accuracy and F1 score. Study quality assessment was performed using the Prediction model of Risk Of Bias Assessment Tool (PROBAST) to assess the risk of bias and concerns regarding applicability.[15]

## Results

Characteristics of included studies

The database searches yielded 10,307 records that met the search criteria, from which 156 were identified as potentially eligible full-text articles (Fig. 1). Ninety studies were excluded as they did not assess rupture risk, 35 were excluded as they investigated correlation of UIA rupture with individual risk factors but did not predict rupture risk, the full texts were not available for 6 studies, and 5 were excluded as they were conference proceedings, non-clinical studies or were retracted. Twenty eligible studies were included (Tables 1 and 2) which spanned from 2003 to 2023, incorporating a total of 20,286 aneurysms[3,6,16-33]. A subset of studies included hold-out test set data with 9/20 studies (45%) only employing training data. All validation was analytical validation (i.e., within a computational research setting), no clinical validation was undertaken (i.e., not within a clinical pathway).[34] The index test model was prospectively validated in 4/20 (20%) studies, and retrospectively validated in 8/20 (40%) studies; the remaining 8/20 (40%) were 'development only' studies. Six studies (6/20, 30%) employed the use of digital subtraction angiography (DSA) or 3D-DSA; 6/20 (30%) used computer tomography angiography (CTA); 3/20 (15%) used a combination of CTA, DSA and magnetic resonance imaging (MRA); and 5/20 (25%) did not specify the imaging modality used. All studies employed as model features a selection of clinical (e.g., age, sex, medical comorbidities), morphological (e.g., size, site, aneurysm or vessel angle, surface area) and/or hemodynamic factors (e.g., wall shear stress, oscillatory shear index, normalized pressure average).[16-33]

**[Figure 1]**

Reference Standards

The reference standard for 16/20 (80%) studies was the occurrence of the stated primary endpoint during follow up: in 11/16 (68.8%) studies this was the occurrence of rupture, and in 5/16 (31.2%) this was 'stability' (the composite primary endpoint of rupture, volumetric growth, and/or presence of symptoms). The duration of follow up was stated in 13/16 (81.3%) of these studies.

The primary reference standard for 3/20 studies (15%) was the UIA risk prediction based arbitrarily on neurosurgeons/neuroradiologist expert opinion. The reference standard for 1/20 (5%) study was the PHASES score for UIA rupture risk prediction.

Index Tests

The index test for each included study involved a form of ML to predict UIA rupture risk or 'stability'. Fourteen (14/20, 70%) studies investigated UIA rupture risk prediction, and 6/20 (30%) aneurysm 'stability'. Some studies investigated the predictive performance of more than one ML model, either independently or as a combination of multiple ML algorithms. The best performing index test (presented in Tables 1 and 2) was a regression or classical ML model in 10/20 (50%) of studies, a deep learning model in 5/20 (25%) of studies, and a combination of ML algorithms in 5/20 (25%) of studies. From the available data, the hold-out test set accuracy range of the regression or classical ML models was 0.67-0.85, deep learning models 0.82-0.85 and combination models 0.66-0.90.

Test Sets

Three studies (3/20, 15%) used geographically separate datasets for training and hold-out testing (external validation) whereas 8/20 (40%) studies used hold-out test sets from the same institution (internal validation). One internal validation study (1/8, 12.5%) used a temporal split. The remaining 9/20 (45%) studies had no hold-out test set and to mitigate overfitting, six of these (6/9, 66.7%) performed either cross validation or bootstrapping.

UIA rupture risk prediction

The overall accuracies of the ML models of the 14/20 (70%) studies that predicted rupture risk ranged between 0.66-0.90 (Table 1), either provided in the publication or calculated by constructing confusion matrices for the purpose of this systematic review.

UIA stability prediction

Of the 6/20 (30%) studies that investigated prediction of UIA 'stability', 3/6 (50%) defined stability as a composite end point of growth or rupture in a specified time period, 2/6 (33.3%) selected a composite end point of growth, rupture or presence of symptoms, and 1/6 (16.7%) did not define stability. The overall accuracies of the ML 'stability' models ranged between 0.83-0.94 (Table 2).

**[Table 1]**

**[Table 2]**

Comparison of index test to other reference standards

As a secondary objective, 6/20 (30%) studies compared the ML model performance to a secondary reference standard – either the prospective predictive performance of either PHASES/UIATS scores or an expert neurosurgeon/neuroradiologist assessment of rupture risk.[22,23,25,30,31,33] As a primary objective, 3/20 (15%) studies compared the ML model performance to the PHASES score or expert assessments, but did not compare the performance to rupture or 'stability'.[16,19,24] ML models had a higher performance accuracy than PHASES/UIATS scores, however, were less accurate than expert clinician predictions. In one study the performance accuracy with PHASES score, the ML model, and expert neurosurgical prediction alone were 0.50, 0.66 and 0.73 AUC-ROC respectively.[25] Another study compared the ML model, expert neurosurgical prediction alone (reader alone) and expert neurosurgical opinion aided by the ML model (reader and model), giving an AUC-ROC of 0.85, 0.88 and 0.95, respectively.[23]

Bias assessment and concerns regarding applicability

An analysis of the risk of bias and concerns regarding applicability was performed for each study using the PROBAST tool (Fig. 2).[15] Two independent reviewers performed the risk of bias assessment, with disagreements resolved by a third reviewer. The PROBAST tool categorizes prediction studies as 'development' or 'validation', which we have assigned as 'training' and 'external validation', respectively, to represent the use of ML in these prediction studies (Note: 'internal validation', whilst reasonably robust when performed temporally in particular, is not considered as 'validation' when explicitly performing PROBAST assessment; nonetheless we include granular information on both internal and external validation in Table 1 and 2). The studies included in this review were all examples of analytical validation (i.e., within a computational research setting).[34] Only 3/20 (15%) studies had geographically separate training sets to their testing sets (external validation).

A 'high' risk of bias or concern for applicability was identified in at least one domain in 13/20 (65%) of studies. Notably, there were high risks of bias in the 'outcome' domain (5/20, 25%) and 'analysis' domain (8/20, 40%). In terms of concerns for applicability in the 'outcome' domain, this was 'unclear' in 11/20 (55%) of studies. These primarily stemmed from studies without a defined duration of follow up to ascertain rupture or 'stability' (assigned as unclear risk), or studies that used subjective reference standards such as expert assessment of rupture risk (assigned as high risk). The inclusion and exclusion criteria were not specified in 7/20 (35%) of studies.

**[Figure 2]**

## Discussion

Summary of studies

The included studies reported high performance accuracies for predicting aneurysm rupture in hold-out tests sets ranging from 0.66-0.90 using ML Models.[16-27] However, the majority of these studies demonstrated a high risk of bias and concerns regarding applicability due to their methodology (Fig. 2), limiting the clinical applicability of their results. Six studies had a composite end-point of 'stability' which included rupture but has debatable clinical applicability. Only three studies used external test sets to validate their performance.[23,26,32] Some studies compared rupture risk prediction to existing clinical methods of prediction, with accuracy of ML models falling between current scoring criteria such as PHASES, and human expert risk prediction.

Current evidence

There is one existing systematic review and meta-analysis for the use of ML in predicting aneurysm rupture risk published in 2022 by Shu, et al.[35] However, only four studies were included and three of these studies predicted current rupture status (i.e., models performing binary classification of aneurysms as 'ruptured' or 'unruptured'), not risk of UIA rupture.[36-38] Therefore, to our knowledge, this is the first review to comprehensively assess studies predicting UIA rupture risk using ML. Other studies not included in this review claim to assess the performance of ML models in predicting UIA rupture risk. However, these studies failed to meet the inclusion criteria for our review as they used the *post-rupture* imaging appearance of ruptured aneurysms to assume pre-rupture morphology, and had no subsequent validation with unruptured cases.[38-43] Furthermore, several longitudinal studies demonstrate that post-rupture morphology is not an adequate substitute to assume pre-rupture morphology when assessing UIA rupture risk as the event of rupture can modify the morphology and haemodynamics of an intracranial aneurysm.[44-46] Epidemiological studies were also excluded if they investigated individual risk factors for rupture but did not develop a predictive model, such as presenting odds ratios for hypertension, aneurysm size or specific morphological features, and aneurysm rupture.[47-49]

The current published validated standards include, for example, the PHASES, UIATS, ISUIA or 'Earlier subarachnoid haemorrhage, aneurysm Location, Age, Population, aneurysm Size and Shape' (ELAPSS) scores to guide management of UIA.[3,5,6,50] The UIATS score was developed by a Delphi consensus, and the ELAPSS score assesses aneurysm growth not rupture, hence neither were included in this review.[5,50] The PHASES score and ISUIA criteria were developed by Cox Proportional Hazard (CPH) analysis, which is based on linear regression modelling.[3-6,51] The PHASES model was developed from a large multicentre cohort of 8,382 aneurysms. Bootstrapping was performed on this training data to account for overfitting and achieved a concordance-statistic value of 0.82 (95% CI 0.79-0.85), however the model was not validated on a hold-out test set (i.e. no internal or external validation).[3] The ISUIA study did not account for overfitting nor were the results validated on a hold-out test set.[6] Nevertheless, models have widely been adopted in practice.[3-6,22,25,52]

Role in clinical practice

The role of aneurysm rupture risk prediction is primarily to guide clinicians and patients towards management options. Decision making requires balancing the risks of intervention (thromboembolic ischaemia, peri-operative haemorrhage, inadequate occlusion, need for re-intervention and death) with the risk of SAH and its subsequent sequelae.[6,53-56] The ISUIA studies reported an overall morbidity and mortality rate for open neurosurgical and endovascular procedures of 7.1%-12.6%, and a meta-analysis of 2460 patients reported a permanent morbidity rate of 10.9%.[6,54] However, it should be noted that these landmark studies were published over two decades ago and significant advances in endovascular interventions have substantially reduced morbidity and mortality.[55,56] A recent meta-analysis of 963 aneurysms treated with an endovascular approach demonstrated a morbidity of 2.85% and mortality of 0.93%.[56] As treatment prediction morbidity and mortality data evolves, so too should rupture risk prediction to allow careful cost-benefit decision making. For now, rupture risk prediction typically involves a pragmatic multi-disciplinary team (MDT) approach incorporating, for example, PHASES, UIATS or ISUIA scores alongside neuroradiological and neurosurgical evaluation of clinical, morphological and hemodynamic factors.[3,5,6] Overall, this complex decision making leads to substantial heterogeneity. Here, novel ML models have the potential to automate and standardize the process of accurately identifying rupture-prone aneurysms. We have shown the high accuracy of ML models, but also have shown that they are not ready for the clinic as higher quality bias-free evidence is needed, as well as clinical validation (i.e., tested within a clinical pathway).

In six studies, the composite end-point of ‘stability’ was defined as a combination of the event of rupture, growth in subsequent imaging and/or presence of symptoms within a specified time period (1-39 months).[28-33] However, several studies have suggested that UIA growth and rupture have a nonlinear and complex relationship.[21,57-59] Therefore, employing stability-based reference standards as a surrogate for risk of rupture may be inaccurate. After all, approximately 3.2% of the general population are expected to have an intracranial aneurysm of any size which almost certainly will have grown at some point in time, but a far fewer proportion of the population experience SAH.[1] Nevertheless, there still remain several studies demonstrating a correlation between growth and rupture which is expected as there is an overlap between the growth risk factors and rupture risk factors.[59,60] A meta-analysis of 4990 aneurysms (13,294 aneurysm-years of follow up) found that growth makes an aneurysm over 30 times more likely to rupture with an annual rupture rate of 3.1% in growing aneurysms.[60] In summary, aneurysm growth is correlated with rupture, and stability can be used as a reference standard for modelling aneurysm risk, but it is not as accurate as rupture being used as the reference standard.

Limitations

There are several limitations regarding the studies included in this review. First, only three studies demonstrated test set data that was geographically separate from training data (i.e., externally validated).[22,26,32] Eight studies used an internal hold-out test set for validation and nine studies had no test set at all. Subsequently there is significant heterogeneity within the methodology of the existing literature. ML algorithms are capable of high accuracy and recall from repetitively being trained on a given data set. Studies without any testing or validation sets are at high risk of overfitting and subsequently have poor external validity. Moreover, with a lack of clinical validation, the generalisability of the results from these studies to the clinic is limited.

Second, as demonstrated by the risk of bias assessment (Fig. 2), several studies demonstrated a high or unclear risk of bias or concerns for applicability. Flaws in the methodology and analysis of several papers, for instance due to ambiguous reference standards, limit the validity of the results predicted. Five studies did not specify a duration of risk assessment or follow-up, such as specifying an annual, 2-year or lifetime risk of rupture. Of these, three studies were also at high risk of observer bias as their reference standard employed was an expert interpretation of UIA rupture risk.[16,19,33] The lack of standardisation between studies make it difficult to appreciate rupture risk across different cohorts. Studies with a short follow-up period are vulnerable to underestimating rupture rates in practice. Conversely, attrition to follow up of stable aneurysms may overestimate rupture risks.

Third, studies investigating prediction of UIA rupture risk are at an inherent risk of selection bias. Naturally, due to ethical concerns of leaving participants untreated when they are at high rupture risk, unstable UIA with known risk factors for rupture are less likely to be longitudinally followed up to a pre-specified endpoint. Consequently, studies excluding high risk unstable UIA that were treated may bias the ML models.[19] Additionally, several studies included specific inclusion and exclusion criteria, such that aneurysms of certain size, location or morphology were excluded. This introduces further risk of selection bias. Furthermore, even subgroup analyses based on such inclusion criteria are challenging because these specific criteria differed amongst the included studies.

Regarding limitations of our systematic review methodology, heterogeneity between study methods, reference standards and testing groups (such as follow-up duration) made it unfeasible to conduct a meta-analysis. Second, the exclusion of pre-prints and non-peer reviewed articles may contribute to publication bias. Due to the mismatch in the pace of peer review compared to developments in data-science, data-science oriented teams may be less inclined to seek publication in peer-reviewed journals than clinically-oriented teams.[61]

<u>Implications for future research and clinical practice</u>

The systematic review serves as a baseline for future research. Despite the potential for homogenous and autonomous data driven decision-making incorporating multiple co-variates, there remains a need for further high-quality studies in this field as ML prediction models are not ready for deployment given the limited evidence, high risks of bias and concerns for applicability. Multicentre, prospectively-maintained international registries of untreated UIAs are needed to train and test ML models. Furthermore, clinical validation (i.e., tested within a clinical pathway) is paramount to establishing the capability of ML in practice. Requirements for clinical validation[34] are to run a prospective clinical study with: (1) a specified end-point (rupture is an unambiguous end-point, and the pre-defined duration would be for as long as study funding would allow); (2) a specified predictive tool (index test) already supported by analytical validation; (3) a clear purpose, including the setting, for which the predictive tool will be used (MDT assessment of intracranial aneurysms); (4) an understanding of the potential benefits and harms associated with that use (a benefit is that rupture risk can be compared to treatment risk in terms of health outcomes allowing informed decisions to be made in either routine clinical practice or for prognostic enrichment in interventional trials; a possible harm is that the predictive tool itself is not entirely proven and therefore may underestimate or overestimate rupture risk; another possible harm is that the predictive tool may lead to uncertainty because there is no risk threshold above or below which a treatment unequivocally should be triggered - therefore treatment discussions remain on a case-to-case basis informed by the risk); (5) a process for collecting and analyzing information about the performance of the predictive tool is devised and carried out (an example would be to include resources in a study to allow frequent follow up and to mitigate drop out by incorporating national epidemiological data collection tools; for example, Hospital Episode Statistics (HES) is a data warehouse containing details of all admissions to NHS hospitals in England). In higher risk aneurysms it is likely that most patients will elect for treatment meaning the recruited cohort will need to be very large.

It is noteworthy that the majority of included studies were developed by training and testing amongst Chinese populations.[16-33] As suggested by the PHASES model, ethnicity plays an integral role in risks of aneurysm rupture.[3] Therefore, studies including ethnicities other than Chinese are also needed. Additionally, ML models could be further developed to meet the needs of individual patients, such as a personalized composite of UIA rupture risk and risk of treatment akin to the UIATS score.[5] Specifically, multi-centric prospective registries (or prospective clinical validation studies) could also include treated patients to elucidate the interaction of treatment. By including data on treated and untreated aneurysms of similar risk profiles under follow up (specified end-points would include morbidity and mortality as well as rupture), counterfactual ML prediction models could be trained to give a prediction risk for treatment and a prediction risk for no treatment.

**Conclusion**

In summary, this systematic review highlights innovative approaches towards rupture risk prediction of UIA using ML models to quantify rupture risk by applying features used in clinical practice. ML algorithms have the potential to automate and standardize the process of accurately identifying rupture-prone aneurysms and determining the need for an inherently high risk prophylactic intervention. However, there remains a need for further high-quality studies in this field as ML prediction models are not ready for deployment given the limited evidence, high risks of bias and concerns for applicability. Further prospective multicentre studies are needed to prove clinical validation before implementation in the clinic.

**Tables**

**Table 1:** Summary of findings for existing studies applying machine learning to predict unruptured intracranial aneurysm rupture risk

| Publication | Study Design | Modality | Reference standard | Time period of risk assessment | Comparison to clinical practice | Index test | Model features | Hold-out test set (n) (or other specified dataset) | Hold-out test set performance accuracy (or performance of other specified dataset) |
|---|---|---|---|---|---|---|---|---|---|
| Weibers et al, 2003[6] | Development only | Not stated | Comparison to result (ruptured/unruptured) | 5 years | - | 1. Linear regression (CPH) | (1) Morphological features | (1692 training set - no hold-out test set) | N/A |
| Grieving et al, 2014[3] | Development only | CTA/DSA/MRA | Comparison to result (ruptured/unruptured) | 5 years | - | 1. Linear regression (CPH) | (1) Clinical features<br>(2) Morphological features | IV: 8328 (Bootstrapping using training data, no hold-out test set per se) | AUC: 0.82 |
| Malik, et al 2018[16] | Development only | DSA | Comparison to risk interpretation of the UIAs was conducted by expert neurosurgeons | Duration not stated | Comparison to neurosurgeon expert opinions for reference standard | 1. MLP Neural network | (1) Morphological features | (12 training set - no hold-out test set) | (Accuracy: 0.86) |
| Jiang et al, 2018[17] | Development and prospective validation | CTA | Comparison to result (ruptured/unruptured) | 18-27 months | - | 1. Logistic regression | (1) Morphological features<br>(2) Hemodynamic features | IV:70 (temporal validation) | Accuracy: 0.67<br>Balanced Accuracy: 0.69<br>Sensitivity: 0.71<br>Specificity: 0.67<br>AUC: 0.72<br>PPV: 0.19<br>NPV: 0.95<br>F1 Score: 0.30 |
| Suzuki et al, 2019[18] | Development and | Not stated | Comparison to result | 2 years | - | 1. Logistic regression* | (1) Clinical features | IV: 102 | Accuracy: 0.82<br>Balanced Accuracy: 0.74 |
| | retrospective validation | | (ruptured/unruptu red) | | | 2. SVM | (2) Morphological features<br>(3) Hemodynamic features | | Sensitivity: 0.64<br>Specificity: 0.85<br>PPV: 0.333<br>NPV: 0.95<br>F1 Score: 0.44 |
| Ahn, et al 2021[19] | Developmen t only | 3D DSA | Comparison to risk interpretation of the UIAs was conducted by two readers, disagreements arbitrated by a third reader. | Duration not stated | Comparison to neuro-radiologist expert opinions for reference standard | 1. Neural network | (1) Morphological features | IV: 93 | Accuracy: 0.82<br>Balanced Accuracy: 0.82<br>Sensitivity: 0.82<br>Specificity: 0.82<br>PPV: 0.80<br>NPV: 0.83<br>F1 Score: 0.81 |
| Ou et al, 2021[20] | Developmen t and retrospective validation | CTA | Comparison to result (ruptured/unruptu red) | 2 years | - | **Combination model:**<br>1. Logistic regression<br>2. LASSO<br>3. Ridge regression | (1) Morphological features<br>(2) Radiomics features | IV: 122 (10-fold cross validation on training data, no hold-out test set per se) | Accuracy: 0.77<br>Balanced Accuracy: 0.80<br>Sensitivity: 0.72<br>Specificity: 0.88<br>AUC: 0.88<br>PPV: 0.92<br>NPV: 0.61<br>F1 Score: 0.81 |
| Van der Kamp, et al 2021[21] | Developmen t only | CTA/DS A/MRA | Comparison to result (ruptured/unruptu red) | 6 months, 1 year, 2 years | - | **Combination model:**<br>1. Linear regression (CPH)<br>2. Ridge regression | (1) Morphological features | (329 training set – no hold-out test set) | (AUC: 0.72) |
| Walther, et al 2022[22] | Developmen t and retrospective validation | Not stated | Comparison to result (ruptured/unruptu red) | Duration not stated | Comparison to PHASES and UIATS | 1. Gradient boosting machine | (1) Clinical features<br>(2) Morphological features | IV: 446 (5-fold cross validation on training data, no hold-out test set per se) | Accuracy: 0.78<br>Balanced Accuracy: 0.78<br>Sensitivity: 0.86<br>Specificity: 0.70<br>AUC: 0.86<br>PPV: 0.74<br>NPV: 0.84<br>F1 Score: 0.80 |
| Ou et al, | Developmen | CTA | Comparison to | 2 years | Comparison to | 1. Neural | (1) Clinical | EV:120 | Accuracy: 0.85 |
| 2022[23] | t and retrospective validation | | result (ruptured/unruptu red) | | individual 5 neurosurgeons aided by AI. (Reader + AI) | network*<br>2. LASSO | features<br>(2) Morphological features | | Balanced Accuracy: 0.87<br>Sensitivity: 0.89<br>Specificity: 0.84<br>AUC: 0.85<br>PPV: 0.69<br>NPV: 0.97<br>F1 Score : 0.88 |
| Wei, et al 2022[24] | Developmen t only | 3D DSA | Comparison made to the PHASES score, patients divided into high and low risk group as reference standard | Duration not stated | Comparison to PHASES for reference standard | 1. Logistic regression | (1) Clinical features<br>(2) Morphological features<br>(3) Hemodynamic features | IV: 39 (Cross validation on training data, no hold-out test set per se) | Accuracy: 0.85<br>Balanced Accuracy: 0.84<br>Sensitivity: 0.78<br>Specificity: 0.90<br>AUC: 0.85<br>PPV: 0.88<br>NPV: 0.83<br>F1 Score: 0.82 |
| Malik, et al 2023[25] | Developmen t and retrospective validation | Not stated | Comparison to result (ruptured/unruptu red) | (i) 5 year and (ii) life time | Comparison to PHASES, UIATS and neurosurgeons expert opinion | **Combination model:**<br>1. Logistic regression<br>2. Decision tree classifier<br>3. Random forest<br>4. AdaBoost<br>5. Gradient boosting<br>6. K-nearest neighbors<br>7. XGBoost. | (1) Clinical features<br>(2) Morphological features | IV: 30 | Accuracy: 0.66<br>Sensitivity: 0.67<br>PPV: 0.73<br>F1-Score: 0.64 |
| Xie, et al 2023[26] | Developmen t and retrospective validation | CTA | Comparison to result (ruptured/unruptu red) | Duration not stated | - | **Combination model:**<br>1. Neural network<br>2. LASSO<br>3. SVM | (1) Clinical features<br>(2) Morphological features<br>(3) Radiomics features | EV: 106 (3-fold cross validation) | Accuracy: 0.90<br>Balanced Accuracy: 0.87<br>Sensitivity: 0.80<br>Specificity: 0.94<br>AUC: 0.89<br>PPV: 0.81<br>NPV: 0.92<br>F1 Score: 0.81 |
| Li, et al 2023[27] | Development only | Not stated | Comparison to result (ruptured/unruptured) | Duration not stated | - | 1. Random forest*<br>2. Logistic regression<br>3. Principal component analysis | (1) Clinical features | IV: 1325 (10-fold cross validation on training data, no hold-out test set per se) | AUC: 0.73 |

Results to 2 decimal places

**Key:**

- *Best Performing ML Model: Where more than one machine learning model was investigated in parallel and not in combination, the best performing model is marked by *.
- AUC: Area under the curve of a receiver operating curve
- LASSO: Least absolute shrinkage and selection operator
- SVM: Support Vector Machine
- MLP: Multilayer Perceptron
- UIA: Unruptured Intracranial aneurysm
- CPH: Cox Proportional Hazards Model
- AI: Artificial Intelligence
- CTA: Computer Tomography Angiogram
- PHASES: Population, Hypertension, Age, Size of aneurysm, Earlier subarachnoid hemorrhage from another aneurysm and Site of aneurysm
- ELAPSS: Earlier subarachnoid hemorrhage, aneurysm Location, Age, Population, aneurysm Size and Shape
- IARS: Intracranial Aneurysm Rupture Score
- UIATS: Unruptured Intracranial Aneurysm Treatment Score
- MRA: Magnetic Resonance Angiogram
- DSA: Digital Subtraction Angiogram
- d.p: decimal places
- IV: Internal Validation
- EV: External Validation
- PPV: Positive Predicitve Value
- NPV: Negative Predictive Value

**Table 2:** Summary of findings for existing studies applying machine learning to predict unruptured intracranial aneurysm 'stability'. The authors defined stability as a composite outcome which included rupture as well as aneurysm growth and/or presence of symptoms.

| Publication | Study Design | Modality | Reference standard | Time period of risk assessment | Definition of stability | Comparison to clinical practice | Index test | Model features | Hold-out test set (n) (or other specified dataset) | Hold-out test set performance accuracy (or performance of other specified dataset) |
|---|---|---|---|---|---|---|---|---|---|---|
| Liu, et al 2019[28] | Development and prospective validation | DSA | Comparison to result (stable/unstable) | 1 month stability assessment (follow-up median: 11.5 months, range: 3–26 months) | 1. Remained unruptured<br>2. No UIA growth<br>3. Asymptomatic | - | 1. Generalized linear model*<br>2. Ridge regression<br>3. Logistic regression | (1) Clinical features<br>(2) Morphological features | IV: 124 | AUC: 0.86 |
| Zhu, et al 2020[29] | Development and retrospective validation | 3D-DSA | Comparison to result (stable/unstable) | 1 month stability assessment (median follow up 15.6 months; range 5–39 months) | 1. Remained unruptured<br>2. No UIA growth | - | 1. Neural Network*<br>2. Random forest<br>3. SVM | (1) Clinical features<br>(2) Morphological features | IV: 411 | Accuracy: 0.82<br>Balanced Accuracy: 0.72<br>Sensitivity: 0.52<br>Specificity: 0.93<br>AUC:0.87<br>PPV: 0.71<br>NPV: 0.85<br>F1 Score: 0.60 |
| Yang, et al 2021[30] | Development and prospective validation | CTA | Comparison to result (stable/unstable) | 3 years | 1. Remained unruptured<br>2. UIA Growth ≤20%. | Comparison made to PHASES, ELAPSS, UIATS and | 1. Neural network | (1) Clinical features<br>(2) Morphological features<br>(3) Hemodynamic | IV: 37 (9-fold cross validation on training data, no | AUC: 0.83 |
| | | | | | | IARS Score | | features | hold-out test set per se) | |
| Liu, et al 2022[31] | Development and prospective validation | CTA | Comparison to result (stable/unstable) | 2 years | 1. Remained unruptured<br>2. UIA of aneurysm <20% or <1mm. | Comparison made to PHASES and ELAPSS | 1. Logistic regression | (1) Clinical features<br>(2) Morphological features<br>(3) Hemodynamic features | IV: 97 | AUC: 0.94 |
| Zhang, et al 2023[32] | Development and retrospective validation | CTA/ MRA | Comparison to result (stable/unstable) | 2 years | 1. Remained unruptured<br>2. No IA growth<br>3. Asymptomatic | - | 1. SVM*<br>2. Logistic regression<br>3. Adaboost | (1) Hemodynamic features | EV: 54 | Accuracy: 0.83<br>Balanced Accuracy: 0.83<br>Sensitivity: 0.83<br>Specificity: 0.83<br>AUC: 0.89<br>PPV: 0.71<br>NPV: 0.91<br>F1 Score: 0.77 |
| Irfan, et al 2023[33] | Development only | DSA | Comparison to risk interpretation of the UIAs was conducted by expert neurosurgeons | Duration not stated | Not defined | Comparison to neurosurgeon expert opinions for reference standard | **Combination model:**<br>1. Neural network<br>2. Decision tree classifier | (1) Morphological features<br>(2) Hemodynamic features | IV: 141 | Accuracy: 0.85<br>Balanced Accuracy: 0.85<br>Sensitivity: 0.84<br>Specificity: 0.86<br>AUC: 0.93<br>PPV: 0.82<br>NPV: 0.87<br>F1 Score: 0.83 |

## Figure Captions

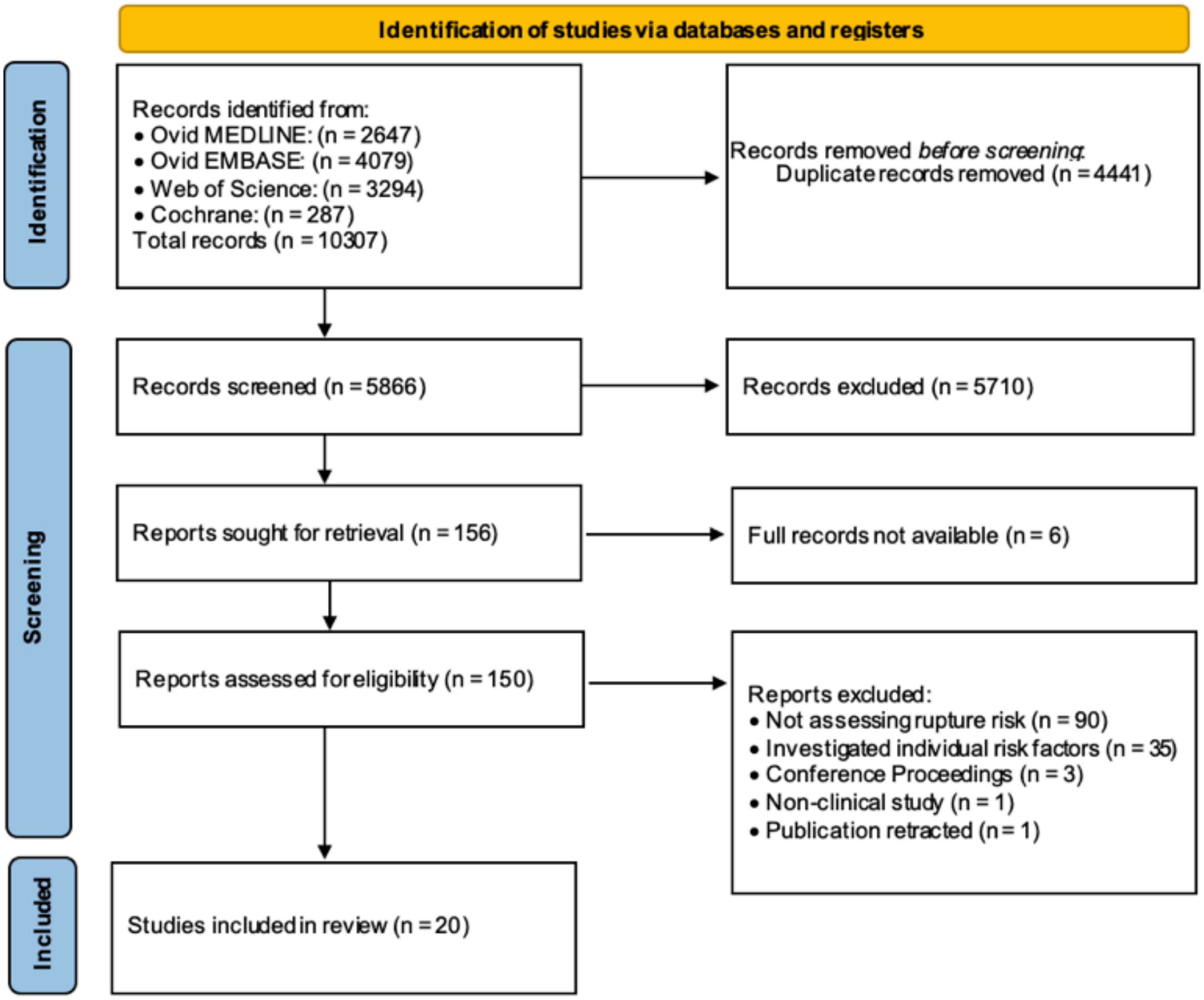

**Figure 1:** PRISMA flow diagram to illustrate the studies included for qualitative review.

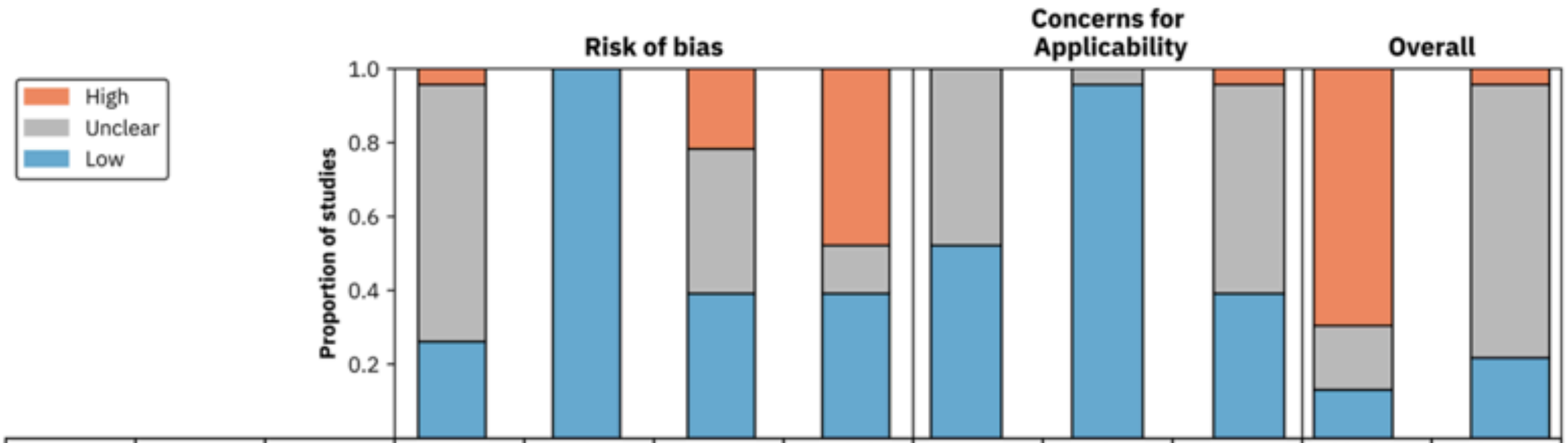

| | | | Risk of bias | | | | Concerns for Applicability | | | Overall | |
|---|---|---|---|---|---|---|---|---|---|---|---|
| **Author** | **Year** | **Validation** | **Participants** | **Predictors** | **Outcome** | **Analysis** | **Participants** | **Predictors** | **Outcome** | **Risk of Bias** | **Applicability** |
| Weibers | 2003 | Internal | Unclear | Low | Unclear | High | Low | Low | Low | High | Unclear |
| Greving | 2014 | Internal | Low | Low | Low | Unclear | Low | Low | Low | Unclear | Low |
| Malik | 2018 | Internal | Unclear | Low | High | High | Unclear | Low | Unclear | High | Unclear |
| Jiang | 2018 | Internal | Unclear | Low | Unclear | Low | Low | Low | Low | Unclear | Low |
| Suzuki | 2019 | Internal | Unclear | Low | Low | Low | Low | Low | Low | Unclear | Unclear |
| Liu | 2019 | Internal | High | Low | Unclear | Low | Unclear | Low | Unclear | High | Unclear |
| Zhu | 2020 | Internal | Low | Low | High | Low | Low | Low | Unclear | High | Unclear |
| Ahn | 2021 | Internal | Unclear | Low | High | Low | Low | Low | Unclear | High | Unclear |
| Ou | 2021 | Internal | Unclear | Low | Low | High | Unclear | Low | Low | High | Unclear |
| Ou | 2021 | External | Unclear | Low | Low | High | Unclear | Low | Low | | |
| van der Kamp | 2021 | Internal | Low | Low | Low | Unclear | Low | Low | Low | Unclear | Low |
| Yang | 2021 | Internal | Unclear | Low | Low | High | Low | Low | Low | High | Low |
| Walther | 2022 | Internal | Unclear | Low | Unclear | High | Unclear | Low | Unclear | High | Unclear |
| Ou | 2022 | Internal | Low | Low | Low | Low | Low | Low | Low | Low | Low |
| Wei | 2022 | Internal | Unclear | Low | Unclear | High | Low | Unclear | Unclear | High | Unclear |
| Liu | 2022 | Internal | Unclear | Low | Unclear | High | Unclear | Low | Unclear | High | Unclear |
| Malik | 2023 | Internal | Unclear | Low | High | Unclear | Unclear | Low | Unclear | High | Unclear |
| Xie | 2023 | Internal | Unclear | Low | Unclear | High | Unclear | Low | Unclear | High | Unclear |
| Xie | 2023 | External | Unclear | Low | Unclear | High | Unclear | Low | Unclear | | |
| Li | 2023 | Internal | Unclear | Low | Unclear | High | Unclear | Low | Unclear | High | Unclear |
| Zhang | 2023 | Internal | Low | Low | Low | Low | Low | Low | Unclear | Low | Unclear |
| Zhang | 2023 | External | Low | Low | Low | Low | Low | Low | Unclear | | |
| Irfan | 2023 | Internal | Unclear | Low | High | Low | Unclear | Low | High | High | High |

**Figure 2:** Summary of the risk of bias and concerns for applicability using PROBAST guidelines explicitly.

**Supplementary Material**

Search strategy - Title & abstract search up till 9th December 2023.

**Ovid EMBASE: 4079 Results**

| Number | Term | Results |
|---|---|---|
| **1** | brain or head or skull or cerebral or cerebrum or intra?cerebral or cranial or cranium or intra?cranial.mp. | 3392753 |
| **2** | exp Brain/ | 1607042 |
| **3** | aneurysm*.mp. | 240601 |
| **4** | exp Aneurysm/ or exp Intracranial Aneurysm | 174770 |
| **5** | ruptur* or burst* or explod* or pop* or breach* or predict* or stability or stable or unstable.mp. | 7777530 |
| **6** | (1 or 2) and (3 or 4) and 5 | 32904 |
| **7** | AI or (artificial adj1 intelligence) or machine learn* or deep learn* or ((deep or machine) adj1 learn*) or neural network* or ((neural or conv*) adj1 (net* or learn* or model*)) or CNN or RNN or convoluted or convnet or computer?assist* or supervised or unsupervised or (semi adj1 supervised).mp. | 453423 |
| **8** | vector machine* or SVM or ((classification or regression or probability or decision) adj1 tree*) or random forest*.mp. | 106167 |
| **9** | ((deep hybrid or cluster* or bayes* or gauss*) adj3 (learn* or model* or net* or algo*)) or algorithm* or automat* or radiomic*.mp. | 1037758 |
| **10** | exp Algorithms/ or exp Artificial Intelligence/ or exp Machine Learning/ or exp Neural Networks, Computer/ or exp Pattern Recognition, Automated/ | 878686 |
| **11** | ((ensemble or transfer or zero shot or reinforcement or dictionary) adj1 (learn* or model* or net* or algo*)).mp. | 26355 |
| **12** | (PCA or principal component analysis or (k adj1 means) or (nearest adj1 neighbo?r) or KNN or (fuzzy adj3 logi*) or isolation forest or hidden markov model or association rule* or feature bag* or score normali#ation).mp. | 196559 |
| **13** | Regression or cox proportional hazard*.mp. | 1726319 |
| **14** | 7 or 8 or 9 or 10 or 11 or 12 or 13 | 3241858 |
| **15** | 6 and 14 | 4079 |

**Ovid Medline: 2647 Results**

| Number | Term | Results |
|---|---|---|
| **1** | brain or head or skull or cerebral or cerebrum or intra?cerebral or cranial or cranium or intra?cranial.mp. | 2495489 |
| **2** | exp Brain/ | 1342435 |
| **3** | aneurysm*.mp. | 180025 |
| **4** | exp Aneurysm/ or exp Intracranial Aneurysm | 135866 |
| **5** | ruptur* or burst* or explod* or pop* or breach* or predict* or stability or stable or unstable.mp. | 5844648 |
| **6** | (1 or 2) and (3 or 4) and 5 | 22330 |
| **7** | AI or (artificial adj1 intelligence) or machine learn* or deep learn* or ((deep or machine) adj1 learn*) or neural network* or ((neural or conv*) adj1 (net* or learn* or model*)) or CNN or RNN or convoluted or convnet or computer?assist* or supervised or unsupervised or (semi adj1 supervised).mp. | 355788 |
| **8** | vector machine* or SVM or ((classification or regression or probability or decision) adj1 tree*) or random forest*.mp. | 76304 |
| **9** | ((deep hybrid or cluster* or bayes* or gauss*) adj3 (learn* or model* or net* or algo*)) or algorithm* or automat* or radiomic*.mp. | 837714 |
| **10** | exp Algorithms/ or exp Artificial Intelligence/ or exp Machine Learning/ or exp Neural Networks, Computer/ or exp Pattern Recognition, Automated/ | 452221 |
| **11** | ((ensemble or transfer or zero shot or reinforcement or dictionary) adj1 (learn* or model* or net* or algo*)).mp. | 23052 |
| **12** | (PCA or principal component analysis or (k adj1 means) or (nearest adj1 neighbo?r) or KNN or (fuzzy adj3 logi*) or isolation forest or hidden markov model or association rule* or feature bag* or score normali#ation).mp. | 136652 |
| **13** | Regression or cox proportional hazard*.mp. | 1189295 |
| **14** | 7 or 8 or 9 or 10 or 11 or 12 or 13 | 2337610 |
| **15** | 6 and 14 | 2647 |

**Web of Science – 3294 results**

| Number | Term | Results |
|---|---|---|
| 1 | ALL=(brain or head or skull or cerebral or cerebrum or intra?cerebral or cranial or cranium or intra?cranial) | 4090366 |
| 2 | ALL=(aneurysm*) | 238452 |
| 3 | ALL=(ruptur* or burst* or explod* or pop* or breach* or predict* or stability or stable or unstable) | 16024526 |
| 4 | #1 AND #2 AND #3 | 24030 |
| 5 | TI=(((artificial NEAR/0 intelligence) or ((deep or machine) NEAR/0 learn*))) OR AB=((artificial NEAR/0 intelligence) or ((deep or machine) NEAR/0 learn*)) | 626428 |
| 6 | ALL=(algorithm* or automat* or radiomic* or computer assist*) | 5149472 |
| 7 | TI = (supervised NEAR/2 (learn* or model* or net* or algo*)) OR AB = (supervised NEAR/2 (learn* or model* or net* or algo*)) | 62528 |
| 8 | TI = (unsupervised NEAR/2 (learn* or model* or net* or algo*) ) OR AB= (unsupervised NEAR/2 (learn* or model* or net* or algo*) ) | 33886 |
| 9 | TI=(semi supervised NEAR/2 (learn* or model* or net* or algo*) ) OR AB=(semi supervised NEAR/2 (learn* or model* or net* or algo*)) | 15617 |
| 10 | TI=(deep hybrid NEAR/2 (learn* or model* or net* or algo*) ) OR AB=(deep hybrid NEAR/2 (learn* or model* or net* or algo*) ) | 16756 |
| 11 | TI=(bayes* NEAR/2 (learn* or model* or net* or algo*) ) OR AB=(bayes* NEAR/2 (learn* or model* or net* or algo*) ) | 92394 |
| 12 | TI=(cluster* NEAR/2 (learn* or model* or net* or algo*) ) OR AB=(cluster* NEAR/2 (learn* or model* or net* or algo*) ) | 124098 |
| 13 | TI=(gauss* NEAR/2 (learn* or model* or net* or algo*) ) OR AB=(gauss* NEAR/2 (learn* or model* or net* or algo*) ) | 57812 |
| 14 | TI = (((neural or conv*) NEAR/0 (net* or learn* or model*) ) or CNN or convnet or RNN) OR AB = (((neural or conv*) NEAR/0 (net* or learn* or model*) ) or CNN or convnet or RNN) | 689702 |
| 15 | TI=(ensemble NEAR/0 (learn* or model* or net* or algo*) ) OR AB=(ensemble NEAR/0 (learn* or model* or net* or algo*) ) | 24783 |
| 16 | TI=(PCA or principal component analysis or (k near/0 means) or (nearest near/0 neighbo$r) or KNN or (fuzzy near/0 logi*) or isolation forest or hidden markov model or association rule* or feature bag* or score normali$ation) OR AB=(PCA or principal component analysis or (k near/0 means) or (nearest near/0 neighbo$r) or KNN or (fuzzy near/0 logi*) or isolation forest or hidden markov model or association rule* or feature bag* or score normali?ation) | 530506 |
| 17 | TI=((vector machine or SVM or ((classification or regression or probability or decision) NEAR/0 tree*) or random forest)) OR AB=((vector machine or SVM or ((classification or regression or probability or decision) NEAR/0 tree*) or random forest)) | 291707 |
| 18 | TI=(regression or cox proportional hazard*) OR AB=(regression or cox proportional hazard*) | 1980400 |
| 19 | #5 OR #6 OR #7 OR #8 OR #9 OR #10 OR #11 OR #12 OR #13 OR #14 OR #15 OR #16 OR #17 or #18 | 8048092 |
| 20 | #4 AND #19 | 3294 |

**Cochrane – 287 results**

| Number | Term | Results |
|---|---|---|
| 1 | brain or head or skull or cerebral or cerebrum or intracerebral or cranial or cranium or intracranial | 146034 |
| 2 | Aneurysm | 4908 |
| 3 | ruptur* or burst* or explod* or pop* or breach* or predict* or stability or stable or unstable | 379008 |
| 4 | MeSH descriptor: [Artificial Intelligence] explode all trees | 2973 |
| 5 | MeSH descriptor: [Neural Networks, Computer] explode all trees | 543 |
| 6 | MeSH descriptor: [Machine Learning] explode all trees | 938 |
| 7 | MeSH descriptor: [Algorithms] explode all trees | 7212 |
| 8 | (artificial NEAR/2 intelligence) or ((deep or machine) NEAR/2 learn*) | 5166 |
| 9 | algorithm or automat* or radiomic* | 33444 |
| 10 | supervised or unsupervised or (semi NEXT supervised) or (deep hybrid or cluster* or bayes* or gauss*) NEAR/2 (learn* or model* or net* or algo*) | 58395 |
| 11 | ((neural or conv*) NEAR/3 (net* or learn* or model*)) or CNN or convnet or RNN or computer?assist* | 14469 |
| 12 | vector machine or SVM or ((classification or regression or probability or decision) NEAR/2 tree*) or random forest* | 9275 |
| 13 | ((ensemble or transfer or zero shot or reinforcement or dictionary) NEAR/2 (learning or model* or net* or algo*)) | 994 |
| 14 | (PCA or principal component analysis or (k NEXT means) or (nearest NEXT neighbo?r) or KNN or (fuzzy NEAR/3 logi*) or isolation forest or hidden markov model or association rule* or feature bag* or score normali?ation) | 14699 |
| 15 | (regression or cox proportional hazard*) | 87380 |
| 16 | #4 or #5 or #6 or #7 or #8 or #9 or #10 or #11 or #12 or #13 or #14 or #15 | 194930 |
| 17 | #1 and #2 and #3 and #16 | 287 |